\documentclass[twoside]{article}
\usepackage{aistats2016_arxiv}

\usepackage{url}
\usepackage{amsthm}

\usepackage{amsmath}
\usepackage{amssymb}
\usepackage{algorithm}
\usepackage{algorithmic}

\newcommand{\ignore}[1]{}
\newcommand{\maxmu}{\mu^{*}}

\newcommand{\KK}{K}

\newcommand{\Indi}[1]{{I\left(#1\right)}}

\newcommand{\G}{L}

\newtheorem{assumption}{Assumption}
\newtheorem{theorem}{Theorem}
\newtheorem{proposition}{Proposition}
\newtheorem*{proof*}{Proof}
\newtheorem{example}{Example}
\newtheorem{corollary}{Corollary}
\newtheorem{definition}{Definition}
\newtheorem{lemma}{Lemma}
\newcommand{\CA}{A}
\newcommand{\GME}{L^{ME}}
\newcommand{\DD}{D}
\newcommand{\UB}{\DD^{UB}}
\newcommand{\FF}{G_*}
\newcommand{\FUB}{U}
\newcommand{\OFF}{G}
\newcommand{\CFF}{F_{*}}
\newcommand{\tT}{t_{T}}
\newcommand{\tS}{t_{F}}

\begin{document}

%

%

\twocolumn[

\aistatstitle{The Max $K$-Armed Bandit: PAC Lower Bounds and Efficient Algorithms}

\aistatsauthor{Yahel David \And Nahum Shimkin}

\aistatsaddress{ Department of Electrical Engineering\\
Technion - Israel Institute of Technology\\
Haifa 32000, Israel
 } 
]

\begin{abstract}
We consider the Max $K$-Armed Bandit problem, where a learning agent is faced with several stochastic arms, each a source of i.i.d. rewards of unknown distribution. 
At each time step the agent chooses an arm, and observes the reward of the obtained sample.  
Each sample is considered here as a separate item with the reward designating its value, and the goal is to find an item with the highest possible value. Our basic assumption is a known lower bound on the {\em tail function} of the reward distributions. Under the PAC framework, we provide a lower bound on the sample complexity of any
$(\epsilon,\delta)$-correct algorithm, and propose an algorithm that attains this bound
up to logarithmic factors.
We analyze the robustness of the proposed algorithm and in addition, we compare the performance of this algorithm to the variant in which the arms
are not distinguishable by the agent and are chosen randomly at each stage.
Interestingly, when the maximal rewards of the arms happen to be similar, the latter
approach may provide better performance.
\end{abstract}

\section{Introduction}
In the classic stochastic multi-armed bandit (MAB) problem the learning agent faces a set $\KK$ of stochastic arms, and wishes to maximize its cumulative reward (in the regret formulation), or find the arm with the highest expected reward (the pure exploration problem). This model has been studied extensively in the statistical and learning literature, see for example \cite{bubeck2012regret} for a comprehensive survey.

We consider a variant of the MAB problem called the Max $K$-Armed Bandit problem (Max-Bandit for short).
In this variant, the objective is to obtain a sample with the highest possible reward (namely, the highest value in the support of the probability distribution of any arm).
More precisely, considering the PAC setting, the objective is to return an $(\epsilon,\delta)$-correct sample,
namely a sample whose value is $\epsilon$-close to the overall best with a probability larger than $1-\delta$. In addition, we wish to minimize the sample complexity, namely the expected number of samples observed by the learning algorithm before it terminates.

For the classical MAB problem, algorithms that find the best arm (in terms of its expected reward) in the PAC sense were presented in \cite{Dar,audibert2010bestArmIdentification,BestArmIdentification_AUnifiedApproach}, and
lower bounds on the sample complexity were presented in \cite{PAC_lower} and \cite{audibert2010bestArmIdentification}. The essential difference with respect to this work is in the objective, which is to find an $(\epsilon,\delta)$-correct sample in our case.
The scenario considered in the Max-Bandit model is relevant when a single best item needs to be selected from
among several (large) clustered sets of items, with each set represented as a single arm.
These sets may represent parts that come from different manufacturers
or produced by different processes, job candidates that are referred by different employment agencies, finding the best match to certain genetic characteristics in different populations,  or choosing the best channel among different frequency bands in a cognitive radio wireless network.

The Max-Bandit problem was apparently first proposed in \cite{cicirello2005max}. For reward distribution functions in a specific family, an algorithm with an upper bound on the sample complexity that increases as $\frac{-\ln(\delta)}{\epsilon^{2}}$ was provided in \cite{streeter2006asymptotically}.
For the case of discrete rewards, another algorithm was presented in \cite{streeter2006simple}, without performance analysis.
Later, a similar model in which the objective is to maximize the expected value of the largest sampled reward for a given number of samples ($n$) was studied in \cite{Extreme_bandits}. In that work the attained best reward is compared with the expected reward obtained by an oracle that samples the best arm $n$ time. An algorithm is suggested and shown to secure an upper bound of order $n^{-b/((b+1)\alpha)}$ on that difference, where $\alpha>0$ and $b>0$ are determined by the properties of the distribution functions and $b$ decreases as they are further away from a specific functions family. Recently, a similar model in which the goal is to find the arm for which the value of a given quantile ($\tau$) is the largest was studied in \cite{DBLP:conf/icml/SzorenyiBWH15}. Their model can be compared to ours by allowing an error $\epsilon$ of the same size as the given quantile.  In this case, the bound on the sample complexity provided in \cite{DBLP:conf/icml/SzorenyiBWH15} increases as $\frac{-\ln\left(\tau\right)-\ln\left(\delta\right)}{\tau^2}$.

Our basic assumption in the present paper is that a known lower bound ($\FF(\epsilon)$, formally defined in Section \ref{sec:model}) is available on the
tail distributions,
namely on the probability that the reward of each given arm will be close to its maximum.\ignore{ However, for the case in which the assumption does not hold, we also provide a robustness analysis of our algorithm.}
A special case is a lower bound on the probability densities near the maximum.
Under that assumption, we provide an algorithm for which the sample complexity increases at most as $\frac{-\ln(\FF(\epsilon)\delta)}{\FF(\epsilon)}$. In the context of \cite{streeter2006asymptotically}, $\FF(\epsilon)\simeq\epsilon$ and in the context of \cite{DBLP:conf/icml/SzorenyiBWH15} $\FF(\epsilon)=\tau$. Therefore, the proposed algorithm provides an improvement by a factor of $\epsilon^{-1}$ over the result of \cite{streeter2006asymptotically}, which was obtained for a more specific model, and an improvement by the same factor over the result of \cite{DBLP:conf/icml/SzorenyiBWH15} which was derived for a similar, but different objective.
To compare with the result in \cite{Extreme_bandits}, we note that by considering the expected maximal value as the maximal possible value, it follows that $\FF(\epsilon)\simeq\epsilon^{\alpha}$. With a choice of $\delta=\frac{1}{n^2}$ in our algorithm, we obtain that the expected deficit of the largest sample with respect to the maximal reward possible is at most of order $O(\frac{\ln(n)}{n^{\alpha}})$ (as compared to $O(n^{-b/((b+1)\alpha)})$ with $b>0$).
Furthermore, we provide a lower bound on the sample complexity of every $(\epsilon,\delta)$-correct algorithm,
which is shown to coincide, up to a logarithmic term, with the upper bound derived for the proposed algorithm. To the best of our knowledge, this is the first lower bound for the present problem.
In addition, we analyze the robustness of the algorithm to our choice of the tail function bound $\FF(\epsilon)$, both for the case where this choice is too optimistic (i.e., the actual distributions do not obey the assumed bound) and for the case where our choice it overly conservative.

A basic feature of the Max-Bandit problem (and the associated algorithms) is the goal of quickly
focusing on the best arm (in term of maximal reward), and sampling from that arm as much as possible.
It is natural to compare the obtained results with an alternative approach, which ignores the
distinction between arms, and simply draws a sample from a random arm at each round. This can be interpreted as mixing the items associated with each arm
before sampling; we accordingly refer to this variant as the unified-arm problem.
This problem actually coincides with the so-called infinitely-many armed bandit model studied in \cite{berry1997bandit,O._Teytaud,infinitely_many_arms_monus,NIPS_Mortal_Multi_Armed_Bandits,NIPS_Two_Target_Algorithms_for_Inf_Armed_Bandit},
for the specific case of deterministic arms studied in \cite{ECML}.
As may be expected, the unified-arm approach provides the best results when the reward distribution of all arms are identical. However, when many arms are suboptimal, the multi-armed approach provides superior performance.

The paper proceeds as follows. In the next section we present our model. In Section \ref{sec:lower} we provide a lower bound on the sample complexity of every $(\epsilon,\delta)$-correct algorithm. In Section \ref{sec:alg} we present an $(\epsilon,\delta)$-correct algorithm, and we provide an upper bound on its sample complexity. The algorithm is simple and its bound has the same order as the lower bound up to a logarithmic term in $\frac{|\KK|}{\epsilon}$ (where $|\KK|$ stands for the number of arms). Then, in Section \ref{sec:robust}, we provide an analysis of the algorithm's performance for the case in which our assumption does not hold. In Section \ref{sec:comp}, we consider for comparison the unified-arm approach. In Section \ref{conc} we close the paper
by some concluding remarks. Certain proofs are differed to the Appendix due to space limitations.

\section{Model Definition}\label{sec:model}

We consider a finite set of arms, denoted by $\KK$. At each stage $t=1,2,\dots$ the learning agent chooses an arm $k\in\KK$, and a real valued reward
is obtained from that arm.
The rewards obtained from each arm $k$ are independent and identically distributed, with a distribution function (CDF) $F_{k}(\mu)$, $\mu\in\mathbb{R}$. We denote the maximal possible reward of each arm by $\mu^{*}_{k}=\inf_{\mu\in \mathbb{R}}\{\mu|F_{k}(\mu)=1\}$, assumed finite, and the maximal reward among all arms by $\maxmu=\max_{k\in\KK} \mu^{*}_{k}$. The tail function $\OFF_k\left(\epsilon\right)$ of each arm is defined as follows.
\begin{definition}
For every arm $k\in\KK$, the tail function $\OFF_k(\epsilon)$ is defined by
\begin{equation*}
\OFF_k(\epsilon)\triangleq 1-F_{k}(\mu^{*}_{k}-\epsilon),\quad \epsilon\geq0\,.
\end{equation*}	
\end{definition}
For example, when $\boldsymbol{\mu}$ is uniform on $[a,b]$,
then $\OFF(\epsilon)=\frac{\epsilon}{b-a}$. In addition, we note that CDFs are nondecreasing functions and therefore the tail functions are non-increasing. It should be observed that $\OFF_k(\epsilon)$ does not reveal the maximal value $\mu^{*}_{k}$, which remains unknown.

Throughout the paper, we shall use the following assumption.
\begin{assumption} \label{assumption:1}
	There exists a known function $\FF(\epsilon)$ and a known constant $\epsilon_0>0$ such that, for every $k\in K$ and $0\leq\epsilon\leq\epsilon_0$, it holds that
	\begin{equation}\label{eq:assumption:1}
	\OFF_k(\epsilon)\geq\FF(\epsilon)\,,
	\end{equation}
\end{assumption}
We note that for every $k\in\KK$, $P\left(\boldsymbol{\mu}_{k}>\mu^{*}_{k}-\epsilon\right)\geq\FF(\epsilon)$ where $\boldsymbol{\mu}_{k}$ stands for a random variable with distribution $F_{k}$. Furthermore, noting that the tail functions are non-negative and non-increasing, we assume the same for their lower bound $\FF(\epsilon)$. Moreover, for convenience we shall assume that $\FF(\epsilon)$ is strictly decreasing in $\epsilon$, and denote its inverse function by $\FF^{-1}(\epsilon)$.

An important special-case is when one assumes that the probability density function (pdf) of each arm is lower bounded by a certain constant $A>0$, so that $\FF(\epsilon)=A\epsilon$. We shall often use the more general bound of the form $\FF(\epsilon)=A\epsilon^{\beta}$ to illustrate our results.

%
%

An algorithm for the Max-Bandit model samples an arm at each time step, based on the observed history so far (i.e., the previously selected arms and observed rewards). We require the algorithm to terminate after a random number $T$ of samples, which is finite with probability 1, and return a reward $V$
which is the maximal reward observed over the entire period.
An algorithm is said to be \emph{$(\epsilon,\delta)$-correct} if
\begin{equation*}
	P\left(V>\maxmu-\epsilon\right)>1-\delta \,.
\end{equation*}
The expected number of samples $E[T]$ taken by the algorithm is the \emph{sample complexity}, which we wish to minimize.

\section{A Lower Bound}\label{sec:lower}

Before turning to our proposed algorithm, we provide a lower bound on the sample complexity of any $(\epsilon,\delta)$-correct algorithm.
The bound is established under Assumption \ref{assumption:1}, and the additional provision that $\FF(\epsilon)$ is concave. The case of non-concave $\FF(\epsilon)$ turns out to be more complicated for analysis, and it is currently unclear whether our lower bound  holds in that case.

For example, when $\FF(\epsilon)=\CA\epsilon^{\beta}$ for some known constants $\CA>0$ and $\beta>1$,
\begin{equation}\label{assum:eps:lower}
P\left(\boldsymbol{\mu}_{k}>\mu^{*}_{k}-\epsilon\right)\geq\CA\epsilon^{\beta}\,,
\end{equation}
the required concavity holds for $\beta\leq1$.
%
The bound in Equation \ref{assum:eps:lower} is usually referred as $\beta$-regularity and is similar to those assumed in \cite{berry1997bandit}, \cite{infinitely_many_arms_monus}, \cite{ECML} and \cite{DBLP:conf/icml/CarpentierV15}.
\ignore{
\begin{assumption}\label{assum2}
	For a given $\epsilon\in(0,\epsilon_{0})$, for every arm $k\in\KK$ for which $\mu^{*}_{k}\geq\maxmu+\epsilon-\epsilon_{0}$, it holds that
	\begin{equation}
		P(\boldsymbol{\mu}_{k}>\mu^{*}_{k}-\Delta_{k}-\epsilon')\geq\CA\left(2\Delta_{k}+\epsilon'\right)^{\beta}\,,
	\end{equation}
	for every $\epsilon'\in[0,\epsilon_{0}-2\Delta_{k}]$, where $\Delta_{k}=\maxmu-\mu^{*}_{k}+\epsilon\leq\frac{\epsilon_{0}}{2}$.
\end{assumption}

It may be verified that if Assumption \ref{assumption:1} holds with a constant $\CA'=\CA2^{\beta}$ (and $\beta>1$),
then Assumption \ref{assum2} holds with the smaller constant $\CA$. In fact we conjecture that Assumption \ref{assum2} is not necessary
for the following lower bound, but a proof is not available at the moment.}

The following result specifies our lower bound.
\begin{theorem}\label{thm:lower:bound}
	Let $k^*$ denote some optimal arm, such that $\mu^{*}_{k^*} = \maxmu$.
	Let Assumption \ref{assumption:1} holds with a concave function $\FF(\epsilon)$ and let $\epsilon\leq\epsilon_0$ and $\delta\leq\frac{3}{20}e^{-3}$. Then, for every $(\epsilon,\delta)$-correct algorithm,
	\begin{equation}\label{eq:lower:bound}
	E[T]\geq \sum_{k\in \KK\setminus \{k^*\}}\frac{1}{32\FF\left(\Theta_k\right)}\ln\left(\frac{3}{20\delta}\right)
	\end{equation}
	where $\Theta_k=\min\left\{\max\left(\epsilon,\maxmu-\mu^{*}_{k}\right),\epsilon_0\right\}$.
\end{theorem}
We note that the specific requirement on $\delta$ is not fundamental, and can be released at the cost of a smaller constant in the bound.

This lower bound can be interpreted as summing over the minimal number of times that each arm, other than the 
optimal arm $k^*$, needs to be sampled. 
It is important to observe that if there are several optimal arms, only one of them is excluded from the summation.
Indeed, the bound is large when there are several optimal (or near-optimal) arms, as the denominator of the summand is small for such arms. This follows since the algorithm needs to obtain more samples to verify that a given arm is $\epsilon$-optimal.

The proof of Theorem \ref{thm:lower:bound} proceeds by considering any given set of reward distributions that obeys the Assumption, and showing that if an algorithm samples some suboptimal arm less than a certain number of times, it cannot be $(\epsilon,\delta)$-correct for some related set of reward distributions for which this arm is optimal.

\emph{Proof of Theorem \ref{thm:lower:bound}.} 
We begin by defining the following set of hypotheses $\{H_0,H_1,\dots,H_{|K|}\}$, where $F^{H_{k}}_{l}(\mu)$ stands for the CDF of arm $l$ under hypothesis $k$ and $\boldsymbol{1}_\Theta$ stands for the indicator function of the set $\Theta$. Hypothesis $H_0$ is the true hypothesis, namely,
\begin{equation*}
F^{H_{0}}_{k}(\mu)=F_{k}(\mu)\quad \forall k\in\KK.
\end{equation*}
For $k=1,\dots,|K|$, we define $H_k$ as follows. For each arm $l\neq k$, its CDF coincides with the true one, namely,
\begin{equation*}
F^{H_{k}}_{l}(\mu)=F_{l}(\mu),\quad l\neq k .
\end{equation*}
For arm $k$, we construct a CDF $F_k^{H_k}$ such that its maximal value is $\mu_k^{*,H_k}=\maxmu+\epsilon$, while it still satisfies Assumption \ref{assumption:1}. 
To define $F^{H_{k}}_{k}$, we use the notation $$\CFF(\mu)=\begin{cases}
1-\FF(\maxmu+\epsilon-\mu) &\mu<\maxmu+\epsilon\\
1 &\mu\geq\maxmu+\epsilon
\end{cases}$$ where $\epsilon$ is provided to the algorithm. We consider two cases.

Case 1: $\mu^{*}_{k}<\maxmu+\epsilon-\epsilon_{0}$. Let
\begin{equation*}
\begin{aligned}
F^{H_{k}}_{k}(\mu)=&\gamma_{k,1}F_{k}(\mu)\boldsymbol{1}_{(-\infty,\mu^{*}_{k})}(\mu)
\\&+\gamma_{k,1}F_{k}(\mu^{*}_{k})\boldsymbol{1}_{[\mu^{*}_{k},\maxmu+\epsilon-\epsilon_{0})}(\mu)
\\&+\CFF(\mu)\boldsymbol{1}_{\left[\maxmu+\epsilon-\epsilon_{0},\infty\right)}(\mu)\,,
\end{aligned}
\end{equation*}
where $\gamma_{k,1}=1-\FF\left(\epsilon_0\right)$. 

Case 2: $\mu^*_{k} \geq  \maxmu +\epsilon- \epsilon_{0} $.
Define $P^{\epsilon}_k\triangleq 1-\FF\left(\epsilon_0\right)+\FF\left(\maxmu+\epsilon-\mu^{*}_{k}\right) \leq 1$, and let
$$\overline{\mu}_{k}=\sup_{\mu\leq \mu^{*}_{k}}\{\mu|F_{k}(\mu)\leq P^{\epsilon}_k\}$$
denote the value for which $F_k$ reaches probability $P^{\epsilon}_k$. Set
\begin{equation*}
\begin{aligned} F^{H_{k}}_{k}(\mu)=&\gamma_{k,2}F_{k}(\mu)\boldsymbol{1}_{(-\infty,\overline{\mu}_{k})}(\mu)
\\&+\left(F_{k}(\mu)+\left(\gamma_{k,2}-1\right)F_{k}(\overline{\mu}_{k})\right)\boldsymbol{1}_{[\overline{\mu}_{k},\mu^{*}_{k})}(\mu)
\\&+\CFF(\mu)\boldsymbol{1}_{\left[\mu^{*}_{k},\infty\right)}(\mu)
\end{aligned}
\end{equation*}
where $\gamma_{k,2}=1-\frac{\FF\left(\maxmu+\epsilon-\mu^{*}_{k}\right)}{F_{k}(\overline{\mu}_{k})}$.

By Lemma \ref{lem:assum}, which is provided in Section \ref{app:lem} in the Appendix, it follows that assumption \ref{assumption:1} holds under all of the hypotheses $\{H_0,H_1,\dots,H_{|K|}\}$.

If hypothesis $H_{k}$ ($k\neq 0$) were true, then $\mu_k^* \geq \mu_l^* +\epsilon$  for all $l\neq k$, hence the algorithm should provide a reward from arm $k$ with probability larger than $1-\delta$. We use $E^{H}_{k}$ and $P^{H}_{k}$ to denote the expectation and probability, respectively, under the algorithm being considered and hypothesis $H_{k}$. For every $k\in\KK$ let
\begin{equation*}
t_{k}=\frac{1}{16\gamma_{k}}\ln\left(\frac{3}{20\delta}\right),
\end{equation*}
where $\gamma_{k}=\FF\left(\epsilon_0\right)$ if $\mu^{*}_{k}< \maxmu+\epsilon-\epsilon_{0}$ and $\gamma_{k}=\FF\left(\maxmu+\epsilon-\mu^{*}_{k}\right)$ if $\mu^{*}_{k}\geq \maxmu+\epsilon-\epsilon_{0}$. In addition, we let $T_{k}$ stand for the number of samples from arm $k$.

Suppose now that our algorithm is $(\epsilon,\delta)$-correct under $H_{0}$, and that $E^{H}_0[T_{k}]\leq t_{k}$ for
some $k\in\KK$. 
We will show that this algorithm cannot be $(\epsilon,\delta)$-correct under hypothesis $H_{k}$. Therefore, an $(\epsilon,\delta)$-correct algorithm must have $E^{H}_0[T_{k}]>t_{k}$ for all $k\in\KK$.

Define the following events , for $k\in \KK$:
\begin{itemize}
	\item $A_{k}=\{T_{k}\leq 4t_{k}\}$. It easily follows from
	$
	4t_{k}\left(1-P^{H}_{0}(A_{k})\right)\leq E^{H}_{0}[T_{k}]
	$
	that if $E^{H}_{0}[T_{k}]\leq t_{k}$, then $P^{H}_{0}(A_{k})\geq\frac{3}{4}$.
	\item
	Let $B_{k}$ stand for the event under which the chosen arm at termination is $k$, and $B^{C}_{k}$ for its complement. Since $P^{H}_{0}\left(B_{k'}\right)>\frac{1}{2}$ can hold for one arm at most, it follows that $P^{H}_{0}\left(B^{C}_{k}\right)>\frac{1}{2}$ for every $k\neq k'$ for some $k'$.
	\item
	Let $C_{k}$ to be the event under which all the samples obtained from arm $k$ are on the interval $(-\infty,\mu^{*}_{k}]$. Clearly, $P^{H}_{0}(C_{k})=1$.
	\item
	For $k\in\KK$ for which $\mu^{*}_{k}< \maxmu+\epsilon-\epsilon_{0}$, $\overline{\mu}_k$ is still defined as before, so $\overline{\mu}_k=\mu^{*}_{k}$ (and $F_k(\overline{\mu}_k)=1$). Now, for every $k\in\KK$, we let $D_k$ denote the event under which for any number of samples $t\leq4t_k$ from arm $k$, the number of samples which are on the interval $(-\infty,\overline{\mu}_k]$ is bounded as follows:
	$$D_k\triangleq\left\{\max_{1\leq t\leq 4t_k}\sum_{i=1}^{t}\left(x^k_i-tF_k\left(\overline{\mu}_k\right)\right)<15t_kF_k\left(\overline{\mu}_k\right)\right\}$$
	where $x^k_i$ is a RV which equals to $1$ if the $i$-th sample from arm $k$ is on that interval and $0$ otherwise. Below we upper bound $P^{H}_{0}\left(D_k\right)$ using Kolmogorov's inequality.
\end{itemize}
Kolmogorov's inequality states that the sum $S_t=\sum_{i=1}^t z_i$ of zero-mean iid random variables $(z_i)$ satisfies $P(\max_{1\leq t\leq n} |S_t|\geq a) \leq \frac{\text{Var}[S_n]}{a^2}$ (Theorem 22.4, in p. 287 of \cite{billingsley1995probability}). By applying it to the RVs $y^{k}_{i}=x^{k}_{i}-F_k\left(\overline{\mu}_k\right)$, we obtain
$$P_0^H\left(D_k^{C}\right)\leq \frac{Var(\sum_{i=1}^{4t_k}y^{k}_{i})}{\left(15t_kF_k\left(\overline{\mu}_k\right)\right)^2}=\frac{4t_kF_k\left(\overline{\mu}_k\right)\left(1-F_k\left(\overline{\mu}_k\right)\right)}{\left(15t_kF_k\left(\overline{\mu}_k\right)\right)^2}\,,$$
where $D_k^{C}$ is the complementary of $D_k$. 

So, for the case of $\mu^{*}_{k}< \maxmu+\epsilon-\epsilon_{0}$, by the fact that $F_k(\overline{\mu}_k)=1$, it follows that $P^{H}_{0}\left(D_k\right)=1$.

For the case of $\mu^{*}_{k}\geq \maxmu+\epsilon-\epsilon_{0}$, it follows that $\FF(\cdot)\leq1$ by its definition, so, again by definition we obtain that $F_k\left(\overline{\mu}_k\right)\geq\FF(\maxmu+\epsilon-\mu)=\gamma_k$ and therefore $t_kF_k\left(\overline{\mu}_k\right)\geq\frac{1}{16}\ln\left(\frac{3}{20\delta}\right)$. So it follows that since $\delta\leq\frac{3}{20}e^{-3}$ by assumption $P^{H}_{0}\left(D_k\right)\geq1-\frac{64}{225\ln\left(\frac{3}{20\delta}\right)}\geq \frac{9}{10}$.
For simplicity, we use the bound $P^{H}_{0}\left(D_k\right)\geq \frac{9}{10}$ for every $k\in\KK$.

Define now the intersection event $S_{k}=A_{k}\cap B_{k}^{C}\cap C_{k}\cap D_{k}$. We have just shown that for every $k\neq k'$ it holds that $P^{H}_{0}(A_{k})\geq\frac{3}{4}$, $P^{H}_{0}(B_{k}^{C})>\frac{1}{2}$, $P^{H}_{0}(C_{k})=1$ and $P^{H}_{0}(D_{k})\geq\frac{9}{10}$, from which it follows that $P^{H}_{0}\left(S_{k}\right)>\frac{3}{20}$ for $k\neq k'$.

Now, we let $h$ to be the history of the process (the sequence of chosen arms and obtained rewards). For every $k\in\KK$, we denote the number of rewards under $\overline{\mu}_k$ by $N_k$. For a given history, at time $t'$, for every $k\in \KK$, the probability of choosing the next arm is the same under $H_0$ and under $H_k$. Also, by the hypotheses definition, the reward probability is the same, unless the chosen arm is $k$. Therefore, by the definition of the hypotheses,
$$\frac{dP^{H}_{k}}{dP^{H}_{0}}(h)=\left(1-\frac{\gamma_{k}}{F_k(\overline{\mu}_k)}\right)^{N_k}f_k(h)$$ where $\gamma_{k}$ is defined before, by the fact that $F_k(\overline{\mu}_k)=1$ for $\mu^{*}_{k}<\maxmu+\epsilon-\epsilon_{0}$, it follows that $\gamma_{k,1}=1-\frac{\gamma_{k}}{F_k(\overline{\mu}_k)}$ for $\mu^{*}_{k}<\maxmu+\epsilon-\epsilon_{0}$ and that $\gamma_{k,2}=1-\frac{\gamma_{k}}{F_k(\overline{\mu}_k)}$ otherwise ($\gamma_{k,1}$ and $\gamma_{k,1}$ are defined before). In addition, $f_k(h)$ represents the Contribution of samples from arm $k$ with rewards strictly larger than $\overline{\mu}_k$.

Now we assume that the intersection event $S_k$ occurs. Then, $C_k$ occurs, so $f_k(h)=1$. Also, $\{A_k\cap D_k\}$ occurs, so $N_k\leq 16t_kF_k(\overline{\mu}_k)$. Therefore, for $\alpha_k=\frac{\gamma_k}{F_k(\overline{\mu}_k)}\leq1$,
$$\frac{dP^{H}_{k}}{dP^{H}_{0}}(h)\Indi{S_{k}}\geq\left(1-\alpha_k\right)^{\frac{1}{\alpha_k}\ln\left(\frac{3}{20\delta}\right)}\Indi{S_{k}}\,.$$
Now, by the fact that $\left(1-\epsilon\right)^{\frac{1}{\epsilon}}\geq e^{-1}$, we obtain the following inequalities,
\begin{align*}
P^{H}_{k}\left(B^{C}_{k}\right) &\geq P^{H}_{k}\left(S_{k}\right)=E^{H}_{0}\left[\frac{dP^{H}_{k}}{dP^{H}_{0}}(h)\Indi{S_{k}}\right]
\\
&\geq E^{H}_{0}\left[\left(1-\alpha_k\right)^{\frac{1}{\alpha_k}\ln\left(\frac{3}{20\delta}\right)}\Indi{S_{k}}\right]
\\&\geq\left(1-\alpha_k\right)^{\frac{1}{\alpha_k}\ln\left(\frac{3}{20\delta}\right)}P^{H}_{0}\left(\Indi{S_{k}}\right)
\\&> \frac{3}{20}e^{-\ln\frac{3}{20\delta}}\geq\delta\,.
\end{align*}
%

We found that if an algorithm is $(\epsilon,\delta)$-correct under hypothesis $H_{0}$ and $E_{0}[T_{k}]\leq t_{k}$ for some $k\neq k'$, then, under hypothesis $H_{k}$ this algorithm returns a sample that is smaller by at least $\epsilon$ than the maximal possible reward with probability of $\delta$ or more, hence the algorithm is not $(\epsilon,\delta)$-correct.
Therefore, any $(\epsilon,\delta)$-correct algorithm must satisfy $E_{0}[T_{k}]> t_{k}$ for all of arms except possibly for one (namely, for the one $k'$ for which $P_{0}\left(B^{C}_{k'}\right)\leq\frac{1}{2}$). In addition $t_{k^{*}}\geq t_{k'}$, where $k^{*}$ is the optimal arm (namely, $\mu^{*}_{k^{*}}=\maxmu$). Hence,
\begin{equation*}
E[T]\geq \sum_{k\in \KK\setminus \{k^*\}}\frac{1}{\FF(\min\left(\epsilon_0,\epsilon+\maxmu-\mu^{*}_{k}\right)}\ln\left(\frac{3}{20\delta}\right).
\end{equation*}
Now, by the fact that $\FF$ is concave, it follows that $t\FF(y)+(1-t)\FF(0)\leq\FF(ty)$ where $y=\maxmu+\epsilon-\mu^{*}_{k}$. So, for the case of $\epsilon\leq\maxmu-\mu^{*}_{k}$, for $t=\frac{\epsilon}{y}$, by the fact that $\FF$ is non-negative, it follows that $\FF(y)\geq2\FF(\epsilon)$ and for the case of $\epsilon<\maxmu-\mu^{*}_{k}$, for $t=\frac{\maxmu-\mu^{*}_{k}}{y}$, it follows that $\FF(y)\geq2\FF(\maxmu-\mu^{*}_{k})$. Then since $\FF$ is a non-decreasing function, the lower bound is obtained.

\qed

\section{Algorithm}\label{sec:alg}
Here we provide an $(\epsilon,\delta)$-correct algorithm. The algorithm is based on sampling the arm which has the highest upper confidence bound on its {\em maximal} reward.

The algorithm starts by sampling a fixed number of times from each arm. Then, it repeatedly calculates an index for each arm which can be interpreted as an upper bound on the maximal reward of this arm, and samples once from the arm with the largest index. The algorithm terminates when the number of samples from the arm with the largest index is above a certain threshold. This idea is similar to that in the UCB1 Algorithm of \cite{Auer}.
\begin{algorithm}[tb]
	\caption{Maximal Confidence Bound (Max-CB) Algorithm}
	\label{alg:example}
	\begin{algorithmic}[1]
		\STATE {\bfseries Input:} The tail function bound $\FF=\{\FF(\epsilon'), 0\le \epsilon' \le \epsilon_0\}$ and its inverse function $\FF^{-1}$, constants $\delta>0$ and $\epsilon>0$.\\ Define $\G=6\ln\left(|\KK|\left(1+\frac{-\ln(\delta)}{\FF\left(\epsilon\right)}\right)\right)$.
		\STATE {\bfseries Initialization:} Counters $C(k)=N_0$, $k\in K$,\\
		where $N_0=\lfloor\frac{\G-\ln{\delta}}{\epsilon_0}\rfloor+1$.
		\STATE Sample $N_0$ times from each arm.\label{begin:mu}
		\STATE Compute $Y^{k}_{C(k)}=V^{k}_{C(k)}+\FUB(C(k))$ and set $k^{*} \in \arg\max_{k\in\KK}Y^{k}_{C(k)}$ (with tie broken arbitrary),
		where $V^{k}_{C(k)}$ is the largest reward observed so far from arm $k$ and $$\FUB(C(k))=\FF^{-1}\left(\frac{\G-\ln(\delta)}{C(k)}\right)\,.$$\label{cond1}
		\STATE If $\FUB(C\left(k^{*}\right))<\epsilon$, stop and return the largest sampled reward.\\
		Else, sample once from arm $k^{*}$, set $C(k^{*})=C(k^{*})+1$ and return to step \ref{cond1}.\label{cond2}
	\end{algorithmic}
\end{algorithm}
\begin{theorem}\label{thm:alg1}
Under Assumption \ref{assumption:1}, for any $\epsilon$ and $\delta$ such that $\G\geq10$, Algorithm \ref{alg:example} is $(\epsilon,\delta)$-correct with a sample complexity of
\begin{equation}\label{eq:thm:main}
E[T]\leq
\sum_{k\in\KK}\frac{\G-\ln(\delta)}{\FF\left(\Theta_k\right)}+|\KK|,
\end{equation}
where $\G=6\ln\left(|\KK|\left(1+\frac{-\ln(\delta)}{\FF\left(\epsilon\right)}\right)\right)$ as defined in the algorithm, and $\Theta_k=\min\left\{\max\left(\epsilon,\maxmu-\mu^{*}_{k}\right),\epsilon_0\right\}$.
\end{theorem}
As observed by comparing the bounds in Equations \eqref{eq:lower:bound} and \eqref{eq:thm:main}, the upper bound in Theorem \ref{thm:alg1} has the same dependence of $\epsilon$ and $\ln(\delta^{-1})$, up to a logarithmic term. It should be noted though that while the lower bound is currently restricted to concave tail function bounds, the algorithm and its bound are not restricted to this case.  

To establish Theorem \ref{thm:alg1}, we first bound the probability of the event under which the upper bound of the best arm is below the maximal reward, using an extreme value bound. Then, we bound the largest number of samples after which the algorithm terminates under the assumption that the upper bound of the best arm is above the maximal reward.

\emph{Proof of Theorem \ref{thm:alg1}}
We denote the time step of the algorithm by $t$, and the value of the counter $C(k)$ at time step $t$ by $C^{t}(k)$. Recall that $T$ stands for the random final time step. By the condition in step \ref{cond2} of the algorithm, for every arm $k\in\KK$, it follows that,
\begin{equation}\label{C_bound}
C^{T}(k)\leq \lfloor\frac{\G-\ln(\delta)}{\FF \left(\epsilon\right)}\rfloor+1.
\end{equation}
Note that by the fact that for $x\geq6$ it follows that $\frac{d6\ln(x)}{dx}\leq1$, and by the fact that for $x_{0}=\exp\left(1\frac{2}{3}\right)$ it follows that $x_{0}>6\ln(x_{0})=10$ it is obtained that
\begin{equation*}
\begin{aligned}
L'&\triangleq|\KK|\left(\frac{-\ln(\delta)}{\FF\left(\epsilon\right)} +1\right)
\\&>6\ln\left(|\KK|\left(\frac{-\ln(\delta)}{\FF\left(\epsilon\right)} +1\right)\right)=\G,
\end{aligned}
\end{equation*}
for $L\geq10$.
So, by the fact that $T=\sum_{k\in\KK}C^{T}(i)$, for $\G\geq10$ it follows that
\begin{equation}\label{time:bound}
\begin{aligned}
T&\leq |\KK|\left(\frac{\G-\ln(\delta)}{\FF\left(\epsilon\right)} +1\right) < |\KK|\left(\frac{L'-\ln(\delta)}{\FF\left(\epsilon\right)} +1\right) 
\\&\leq L'^{2}=e^{\frac{\G}{3}}.
\end{aligned}
\end{equation}
Now, we begin with proving the $(\epsilon,\delta)$-correctness property of the algorithm.
Recall that for every arm $k\in\KK$ the rewards are distributed according to the CDF $F_{k}(\mu)$. Let assume w.l.o.g. that $\mu^{*}_{1}=\maxmu$. Then, for $N>0$ and by the fact that $(1-\epsilon)^\frac{1}{\epsilon}\leq e^{-1}$ for every $\epsilon\in(0,1]$, for $\FUB(N)=\FF^{-1}\left(\frac{\G-\ln(\delta)}{N}\right)$ it follows that
\begin{equation}\label{eq:delta:bound:pre}
\begin{aligned}
P\left(V^{1}_{N}\leq\maxmu-\FUB(N)\right)&=\left(F_{1}\left(\maxmu-\FUB(N)\right)\right)^{N}
\\&\leq \left(1-\left(\frac{\G-\ln(\delta)}{N}\right)\right)^{N}
\\&\leq \delta e^{-L},
\end{aligned}
\end{equation}
where $V^{k}_{N}$ is the largest reward observed from arm $k\in\KK$ after this arm has been sampled for $N$ times. Hence, at every time step $t$, by the definition of $Y^{1}_{C^{t}(1)}$ and Equations \eqref{time:bound} and \eqref{eq:delta:bound:pre}, by applying the union bound, it follows that
\begin{equation}\label{eq:delta:bound}
\begin{aligned}
P\left(Y^{1}_{C^{t}(1)}\leq\maxmu\right)&= P\left(V^{1}_{C^{t}(1)}\leq\maxmu-\FUB(C^{t}(1))\right)
\\&\leq \sum_{t=1}^{\exp\left(\frac{\G}{3}\right)}P\left(V^{1}_{N}
\leq\maxmu-\FUB(N)\right)
\\&\leq \delta e^{-\frac{2L}{3}}.
\end{aligned}
\end{equation}

Since by the condition in step \ref{cond2}, it is obtained that when the algorithm stops
\begin{equation*}
V^{k^{*}}_{C^{t}(k^{*})}>Y^{k^{*}}_{C^{t}(k^{*})}-\epsilon,
\end{equation*}
and by the fact that for every time step
\begin{equation*}
Y^{k^{*}}_{C^{t}(k^{*})}\geq Y^{1}_{C^{t}(1)},
\end{equation*}
it follows by Equation \eqref{eq:delta:bound} that
\begin{equation*}
P\left(V^{k^{*}}_{C^{t}(k^{*})}\leq\maxmu-\epsilon\right)\leq P\left(Y^{1}_{C^{t}(1)}\leq\maxmu\right)\leq\delta e^{-\frac{2L}{3}}.
\end{equation*}
Therefore, it follows that the algorithm returns a reward greater than $\maxmu-\epsilon$ with a probability larger than $1-\delta$. So, it is $(\epsilon,\delta)$-correct.\\

For proving the bound on the expected sample complexity of the algorithm we define the following sets:
$$M(\epsilon)=\{l\in\KK|\maxmu-\mu^{*}_{l}<\epsilon\}$$
and $$N(\epsilon)=\{l\in\KK|\maxmu-\mu^{*}_{l}\geq\epsilon\}.$$
As before, we assume w.l.o.g. that $\mu^{*}_{1}=\maxmu$.
For the case in which
\begin{equation*}
E_{1}\triangleq\bigcap_{1\leq t<T}\left\{Y^{1}_{C^{t}(1)}\geq\maxmu\right\},
\end{equation*}
occurs, since $V^{k}_{C^{t}(k)}\leq\mu^{*}_{k}$ for every $k\in\KK$, and every time step, it follows that the necessary condition for sampling from arm $k$,
\begin{equation*}
Y^{k}_{C^{k}(1)}\geq Y^{1}_{C^{t}(1)},
\end{equation*}
occurs only when the event
\begin{equation*}
E_{2}(t)\triangleq\left\{\mu^{*}_{k}+\FUB\left(C^{t}(k)\right)\geq \maxmu\right\},
\end{equation*}
occurs. But
\begin{equation*}
E_{2}(t)\subseteq \left\{C^{t}(k)\leq \frac{\G-\ln(\delta)}{\FF\left(\maxmu-\mu^{*}_{k}\right)}\right\}.
\end{equation*}
Therefore, it is obtained that
\begin{equation}\label{bound_set_N}
C^{T}(k)\leq \max\left(\lfloor\frac{\G-\ln(\delta)}{\FF\left(\maxmu-\mu^{*}_{k}\right)}\rfloor+1,N_0\right).
\end{equation}
By using the bound in Equation \eqref{C_bound} for the arms in the set $M(\epsilon)$, the bound in Equation \eqref{bound_set_N} for the arms in the set $N(\epsilon)$ and the bound in Equation \eqref{time:bound}, it is obtained that
\begin{equation}\label{eq:final:1}
E[T]\leq \left(1-P\left(E_{1}\right)\right)e^{\frac{\G}{3}}+P\left(E_{1}\right)\Phi\left(\epsilon\right),
\end{equation}
where
\begin{equation*}
\begin{aligned}
\Phi\left(\epsilon\right)\triangleq&\sum_{k\in N(\epsilon)}\left(\lfloor\frac{\G-\ln(\delta)}{\FF\left(\min\left(\epsilon_0,\maxmu-\mu^{*}_{k}\right)\right)}\rfloor+1\right)
\\&+\sum_{k\in M(\epsilon)}\left(\lfloor\frac{\G-\ln(\delta)}{\FF\left(\epsilon\right)}\rfloor+1\right).
\end{aligned}
\end{equation*}
In addition, by Equation \eqref{eq:delta:bound}, the bound in Equation \eqref{time:bound} and by applying the union bound, it follows that
\begin{equation*}
\begin{aligned}
P\left(E_{1}\right)\geq1-\sum_{t=1}^{T}P\left(Y^{1}_{C^{t}(1)}<\maxmu\right)&\geq 1-\delta e^{-\frac{2L}{3}}e^{\frac{L}{3}}
\\&=1-\delta e^{-\frac{L}{3}}.
\end{aligned}
\end{equation*}
So,
\begin{equation}\label{eq:final:2}
1-P\left(E_{1}\right)\leq\delta e^{-\frac{L}{3}}.
\end{equation}
Furthermore, by the definitions of the sets $N(\epsilon)$ and $M(\epsilon)$ and since $\epsilon\leq\epsilon_0$, it can be obtained that
\begin{equation}\label{eq:final:3}
\Phi\left(\epsilon\right)\leq\sum_{k\in \KK}\lfloor\frac{\G-\ln(\delta)}{\FF\left(\Theta_k\right)}\rfloor+1.
\end{equation}
where $\Theta_k=\min\left\{\max\left(\epsilon,\maxmu-\mu^{*}_{k}\right),\epsilon_0\right\}$.
Therefore, by Equation \eqref{eq:final:1}, \eqref{eq:final:2} and \eqref{eq:final:3} the bound on the sample complexity is obtained. 

\qed

\section{Robustness}\label{sec:robust}
The performance bounds presented for our algorithm depend directly on the choice of the lower bound $\FF$ on the tail functions.  A natural question is what happens if our choice of $\FF$ is too optimistic, so that  Assumption 1 is violated.  In the opposite direction, how tight is our bound when our choice of $\FF$ is to conservative?  We address these two questions in turn. 
\subsection{Optimistic Tails Estimate}
Here Equation \eqref{eq:assumption:1} does not hold for $\FF(\epsilon)$, but holds for $\FF'(\epsilon)=\alpha\FF(\epsilon)$ for some $\alpha<1$.
The fact that Equation \eqref{eq:assumption:1} does not hold for $\FF(\epsilon)$ leads to the situation in which the probability $P\left(Y^{k}_{C(k)}<\mu^{*}_{k}\right)$ is larger (where $Y^{k}_{C(k)}$ is the index calculated in step \ref{cond1} of the algorithm) than the value on which the proof of Theorem \ref{thm:alg1} relies. In the following proposition we provide the $(\epsilon,\delta)$-correctness and sample complexity of Algorithm \ref{alg:example}.
\begin{proposition}\label{prop:1}
	Suppose that Assumption \ref{assumption:1} does not hold for $\FF(\epsilon)$, but holds for $\FF'(\epsilon)=\alpha\FF(\epsilon)$ for some $\alpha<1$. Then Algorithm \ref{alg:example} is $(\epsilon',\delta')$-correct with $$\epsilon'=\FF^{-1}\left(\left(|\KK|\left(\G-\ln(\delta)\right)\right)^{1-\alpha}\left(\FF\left(\epsilon\right)\right)^{\alpha}\right)\, \text{and}\,\, \delta'=\delta^{\alpha},$$ and sample complexity bound
	\begin{equation*}
	\begin{aligned}
	E[T]\leq
	&\left(1-\delta^{\alpha}\right)\sum_{k\in\KK}\frac{\G-\ln(\delta)}{\FF\left(\overline{\Theta}_k\right)}
	\\&+\delta^{\alpha}\left(|\KK|\left(\frac{\G-\ln(\delta)}{\FF\left(\epsilon\right)}\right)\right)^{1-\alpha}+|\KK|,
	\end{aligned}
	\end{equation*}
	where $L$ is as in Theorem \ref{thm:alg1} and $\overline{\Theta}_k=\min\left\{\max\left(\epsilon,\maxmu-\epsilon'-\mu^{*}_{k}\right),\epsilon_0\right\}$.
\end{proposition}
Note that for $\alpha=1$, the bound of Theorem \ref{thm:alg1} is recovered. The proof of the above proposition bases on the proof of Theorem \ref{thm:alg1} and is provided in Section \ref{sec:proof:prop:1} in the Appendix.

\subsection{Conservative Tails Estimate}
Here, Assumption \ref{assumption:1} holds for the provided function $\FF(\epsilon)$  and also holds for $\FF'(\epsilon)=\alpha\FF(\epsilon)$ for some $\alpha>1$. Therefore, in this case the probability $P\left(Y^{k}_{C(k)}<\mu^{*}_{k}\right)$ is smaller than the value on which the proof of Theorem \ref{thm:alg1} relies. So, Algorithm \ref{alg:example} returns an $\epsilon$-optimal value with a larger probability. The probability of returning a false value is given in the following proposition.
\ignore{
\begin{proposition}\label{prop:2e}
	When Assumption \ref{assumption:1} holds for $\FF(\epsilon)$, and also for $\FF'(\epsilon)=\alpha\FF(\epsilon)$ for some $\alpha>1$, and $\G\geq10$, Algorithm \ref{alg:example} is $(\epsilon-\epsilon'',\delta')$-correct where $$\epsilon''=\min_{N_0\leq x\leq e^{\G/3}}\left(\FUB\left(x\right)-\FUB\left(\frac{x}{\alpha}\right)\right)\,\text{and}\,\, \delta'=\delta^{\alpha},$$ ($\epsilon$ and $\delta$ are provided to the algorithm and $N_0$ is defined in the algorithm) with a sample complexity of
\begin{equation*}
E[T]\leq
\sum_{k\in\KK}\frac{\G-\ln(\delta)}{\max\left(\FF\left(\epsilon\right),\FF\left(\maxmu+\epsilon''-\mu^{*}_{k}\right)\right)}+|\KK|,
\end{equation*}
	where $\G=6\ln\left(|\KK|\left(1+\frac{-\ln(\delta)}{\FF\left(\epsilon\right)}\right)\right)$ as defined in the algorithm.
\end{proposition}
For proving the above proposition we base on a minor variation of the proof of Theorem \ref{thm:alg1}.
\begin{proof*}[Proposition \ref{prop:2e}]
	\qed
\end{proof*}}

\begin{proposition}\label{prop:2}
	When Assumption \ref{assumption:1} holds for $\FF(\epsilon)$, and also for $\FF'(\epsilon)=\alpha\FF(\epsilon)$ for some $\alpha>1$, and $\G\geq10$, Algorithm \ref{alg:example} is $(\epsilon,\delta')$-correct where $\delta'=\delta^{\alpha} e^{-(\alpha-1)L}$ ($\epsilon$ and $\delta$ are provided to the algorithm) with the sample complexity provided in Theorem \ref{thm:alg1}
\end{proposition}
For proving the above proposition we base on a minor variation of the proof of Theorem \ref{thm:alg1}. The proof is provided in Section \ref{proof:prop:2} in the Appendix.

\section{Comparison with the Unified-Arm Model}\label{sec:comp}
In this section, we analyze the improvement in the sample complexity obtained by utilizing the multi arm framework (the ability to choose from which arm to sample at each time step) compared to a model in which all the arms are unified into a single arm, so that the sample is effectively obtained from a random arm. 
In the unified-arm model, when the agent samples from this unified arm, one of the original arms is chosen uniformly at random, and a reward is sampled from this arm. The CDF of the unified arm is therefore $F(\mu)=\frac{1}{|K|} \sum_{k\in \KK} F_k(\mu)$, and the corresponding maximal reward is $\mu^*=\max_k\mu^*_k$. Assumption \ref{assumption:1},  implies that $1-F(\mu)\geq\frac{\FF\left(\maxmu-\mu\right)}{|\KK|}$.


In the remainder of this section, we provide a lower bound on the sample complexity and an $(\epsilon,\delta)$-correct algorithm that attains the same order of this bound for the unified-arm model. (Note that the 
lower bound in Theorem \ref{thm:lower:bound} is meaningless for $|K|=1$.)
Then, we discuss which approach (multi-armed or unified-arm) is better for different model parameters, 
and provide examples that illustrate these cases.

\subsection{Lower Bound}
The following Theorem provides a lower bound on the sample complexity for the unified-arm model.
\begin{theorem}\label{thm:lower:single}
For every $(\epsilon,\delta)$-correct algorithm, under Assumption \ref{assumption:1}, when $\FF(\epsilon)$ is concave and $\delta\leq\frac{3}{20}e^{-3}$, it holds that
\begin{equation}
E[T]\geq \frac{|\KK|}{16\FF\left(\epsilon\right)}\ln\left(\frac{3}{20\delta}\right).
\end{equation}
\end{theorem}
The proof is provided in Section \ref{app:proof:lower:sing} in the Appendix and is based on a similar idea to that of Theorem \ref{thm:lower:bound}.

\subsection{Algorithm}
In Algorithm \ref{alg:example_S}, a fixed number of instances is sampled, and the algorithm chooses the best one among them. In the following Theorem we provide a bound on the sample complexity achieved by Algorithm \ref{alg:example_S}.
\begin{algorithm}[tb]
	\caption{Unified-Arm Algorithm}
	\label{alg:example_S}
	\begin{algorithmic}[1]
		\STATE {\bfseries Input:} The tail function bound $\FF=\{\FF(\epsilon'), 0\le \epsilon' \le \epsilon_0\}$ and its inverse function $\FF^{-1}$, constants $\delta>0$ and $\epsilon>0$.
		\STATE Sample $\lceil\frac{-\ln(\delta)|\KK|}{\FF(\epsilon)}\rceil+1$ times from the unified-arm.\label{begin:mu_S}
		\STATE Return the best sample.
	\end{algorithmic}
\end{algorithm}

\begin{theorem}\label{thm:alg:Single}
Under Assumption \ref{assumption:1}, Algorithm \ref{alg:example_S} is $(\epsilon,\delta)$-correct, with a sample complexity bound of
\begin{equation*}
E[T]\leq
\frac{|\KK|\ln(\delta^{-1})}{\FF(\epsilon)}+2.
\end{equation*}
\end{theorem}
The proof is provided in Section \ref{app:proof:of:al:2} in the Appendix. Note that the upper bound on the sample complexity is of the same order as the lower bound in Theorem \ref{thm:lower:single}.

\subsection{Comparison and Examples}
 To find when the multi-armed algorithm is useful, we may compare the upper bound on the sample complexity provided in Theorem \ref{thm:alg1} for Algorithm \ref{alg:example} (multi-armed case) with the lower bound for the unified-arm model in Theorem \ref{thm:lower:single}. We consider two extreme cases. 
 
{\em Case 1:} Suppose that arm 1 is best: $\mu_1^*=\maxmu$, while all the other arms fall short significantly 
compared to the required accuracy $\epsilon$: $\mu_k^*  \ll \maxmu - \epsilon$, for $k\neq 1$.
\\Here $\frac{1}{\epsilon}\gg\frac{1}{\left(\max\left(\epsilon,\maxmu-\mu^{*}_{k}\right)\right)}$, for $k\neq 1$. Hence the upper bound on sample complexity of Algorithm \ref{alg:example} (multi-armed case) will be smaller than the lower bound for the unified-arm model in Theorem \ref{thm:lower:single}. We now provide an example which illustrates case 1 numerically.
\begin{example}[Case 1]\label{ex:1}
	Let $|\KK|=10^4$, $\mu^{*}_{1}=0.9$, $\mu^{*}_{k}=0.1\ \forall k\neq1$, $\FF(\epsilon)=\CA\epsilon$ and $\CA=0.01$. 
 	For $\epsilon=10^{-4}$ and $\delta=10^{-3}$ the sample complexity attained by Algorithm \ref{alg:example} is $3.52\times10^8$.  The lower bound for the unified-arm model is $3.13\times10^{9}$. The sample complexity attained by Algorithm \ref{alg:example_S} (for the unified-arm model model) is $6.9\times10^{10}$.
\end{example}
{\em Case 2:} Consider next the opposite case, where there are many optimal arms and few that are worse: 
say $\mu_1^* \ll \maxmu -\epsilon$, while $\mu_k^*=\maxmu$ for all $k\neq 1$.
\\Here $\frac{1}{\epsilon}=\frac{1}{\left(\max\left(\epsilon,\maxmu-\mu^{*}_{k}\right)\right)}$, for $k\neq 1$. Hence, since there is a logarithmic-in-$\frac{|\KK|}{\epsilon}$ multiplicative factor in the upper bound on the sample complexity of Algorithm \ref{alg:example}, this bound will be larger than the lower bound for the unified-arm model in Theorem \ref{thm:lower:single}. The following example illustrates case 2 numerically.
\begin{example}[Case 2]\label{example:2}
 	Let $|\KK|$, $\FF(\epsilon)$, $\delta$ and $\epsilon$ remain the same as in Example \ref{ex:1}, and let $\mu^{*}_{1}=0.1$ and $\mu^{*}_{k}=0.9$ for $k\neq1$. The sample complexity of Algorithm \ref{alg:example} is $1.56\times10^{12}$, which is larger than the sample complexity of Algorithm \ref{alg:example_S} which is $6.9\times10^{10}$.
\end{example}

As shown in Example \ref{example:2}, in some cases the bound on the sample complexity of the multi-armed Algorithm \ref{alg:example} is larger than that of the unified-arm Algorithm \ref{alg:example_S}. We shall further comment on these finding in our concluding remarks. 



\section{Conclusion}\label{conc}
We have considered in this paper the Max $K$-armed Bandit problem in the PAC setting, under the assumption of a known lower bound on the tail function of reward distributions. We provided a lower bound on the sample complexity of any algorithm, and a UCB-type sampling algorithm whose sample complexity is essentially of the same order up to logarithmic terms. 

We have further analyzed the robustness of our algorithm to the violation of Assumption \ref{assumption:1} on the tail functions, and bounded the result deterioration in performance which is shown to be gradual.

The performance of the multi-armed Algorithm \ref{alg:example} was compared to a simple unified-arm approach.  The benefits of Algorithm \ref{alg:example}, which aims to focus sampling on the best arms, are clear when there are few optimal arms (in term of their maximal reward), but might diminish when many arms are close to optimal. Combining these two approaches into a single algorithm that excels in either case remains a challenge for future works.      

%
%
%

\vskip 0.2in
\newpage
\bibliography{sample_new2}
\bibliographystyle{abbrv} 
\newpage
\section{Appendix}

\subsection{Lemma \ref{lem:assum}}\label{app:lem}
\begin{lemma}\label{lem:assum}
Assumption \ref{assumption:1} holds under the Hypotheses $\{H_0,H_1,\dots,H_{|K|}\}$ defined in the proof of Theorem \ref{thm:lower:bound}.
\end{lemma}
\proof
For the hypothesis $H_0$ the assumption holds since that is the true one.

For $k\in\KK$ for which $\mu^{*}_{k}<\maxmu+\epsilon-\epsilon_{0}$, since the CDF is $\CFF(\mu)$ on the interval $[\maxmu+\epsilon-\epsilon_0,\mu^{*,H_k}_k]$, which is of size $\epsilon_0$ and reaches the maximal value, it is easily obtained by the definition of $\CFF(\mu)$ that Assumption \ref{assumption:1} holds.

Now we will show that Assumption \ref{assumption:1} holds in the last case, $k\in\KK$ for which $\mu^{*}_{k}<\maxmu+\epsilon-\epsilon_{0}$. We need to show that 
\begin{equation}\label{eq:to:show}
1-F^{H_{k}}_{k}(\maxmu+\epsilon-\epsilon')\geq \FF(\epsilon')
\end{equation}
for every $0\leq\epsilon'\leq\epsilon_0$. For $0\leq\epsilon'\leq \maxmu+\epsilon-\mu^{*}_{k}$, Equation \eqref{eq:to:show} holds by the definition of $\CFF$. For $\maxmu+\epsilon-\overline{\mu}_{k}<\epsilon'\leq\epsilon_0$, by the definition of $\overline{\mu}_{k}$ it follows that $F_{k}(\mu)\leq P^{\epsilon}_{k}$ for $\mu<\overline{\mu}_{k}$, so $F^{H_{k}}_{k}(\maxmu+\epsilon-\epsilon')\leq1-\FF(\epsilon_0)$ and Equation \eqref{eq:to:show} follows by the monotonicity of $\FF(\cdot)$.

Finally, for the case of $\maxmu+\epsilon-\mu^{*}_{k}<\epsilon'\leq \maxmu+\epsilon-\overline{\mu}_{k}$ we use the concavity of $\FF$. For $\overline{\mu}_k\leq\mu<\mu^{*}_{k}$ it follows that $F^{H_{k}}_{k}(\mu)=F_{k}(\mu)-\FF(\maxmu+\epsilon-\mu^{*}_{k})$. Also, by the fact that Assumption \ref{assumption:1} holds for $F_k$ we have that  $\FF(\mu^{*}_{k}-\mu)\leq 1-F_{k}(\mu)$, so that
\begin{equation}\label{eq:F:D}
F^{H_{k}}_{k}(\mu)\leq1-\Delta_{k}
\end{equation}
where $\Delta_{k}=\FF(\mu^{*}_{k}-\mu)+\FF(\maxmu+\epsilon-\mu^{*}_{k})$. Then by the assumed concavity of $\FF$, and noting that $\FF(0)=0$, it follows that $\lambda\FF(\maxmu+\epsilon-\mu)\leq\FF(\lambda\left(\maxmu+\epsilon-\mu\right))$ for $\lambda\in[0,1]$. So by choosing $\lambda=\lambda_1=\frac{\maxmu+\epsilon-\mu^{*}_{k}}{\maxmu+\epsilon-\mu}$ it follows that $\lambda_1\FF(\maxmu+\epsilon-\mu)\leq\FF\left(\maxmu+\epsilon-\mu^{*}_{k}\right)$
and by choosing $\lambda=1-\lambda_1$ it follows that
$(1-\lambda_1)\FF(\maxmu+\epsilon-\mu)\leq\FF\left(\mu^{*}_{k}-\mu\right)$. Therefore,
\begin{equation}\label{eq:con:1}
\FF(\maxmu+\epsilon-\mu)\leq\FF\left(\mu^{*}_{k}-\mu\right)+\FF\left(\maxmu+\epsilon-\mu^{*}_{k}\right)=\Delta_{k}
\end{equation}
So, by Equations \eqref{eq:F:D} and \eqref{eq:con:1} it follows that
\begin{equation*}
F^{H_{k}}_{k}(\mu)\leq1-\FF(\maxmu+\epsilon-\mu)\,,
\end{equation*}
so that Assumption \ref{assumption:1} holds.

\qed

\subsection{Proof of Proposition \ref{prop:1}}\label{sec:proof:prop:1}
\proof
	First we denote the inverse function of $\FF'(\epsilon)$ (the function for which Assumption \ref{assumption:1} holds) by $\FF'^{-1}$, and we note that,
	\begin{equation}\label{prop1:basic}
	\FF'^{-1}(y)=\FF^{-1}\left(\frac{y}{\alpha}\right)\, .
	\end{equation}
	Now, we begin with providing the $(\epsilon',\delta')$-correctness of the algorithm. Let assume w.l.o.g. that $\mu^{*}_{1}=\maxmu$. Then, by Equation \eqref{prop1:basic} it follows that $\FUB(N)=\FF'^{-1}\left(\frac{\alpha\left(\G-\ln(\delta)\right)}{N}\right)$, and therefore similarly to Equation \eqref{eq:delta:bound:pre} it follows that
	\begin{equation}\label{eq:prp1:delta:bound:pre}
	\begin{aligned}
	P\left(V^{1}_{N}\leq\maxmu-\FUB(N)\right)&=\left(F_{1}\left(\maxmu-\FUB(N)\right)\right)^{N}
	\\&\leq \left(1-\left(\frac{\alpha\left(\G-\ln(\delta)\right)}{N}\right)\right)^{N}
	\\&\leq \delta^{\alpha} e^{-\alpha \G},
	\end{aligned}
	\end{equation}
	Hence, for every time step $t\leq \tT$ where $\tT=e^{\alpha\G/3}$, by applying the union bound, the definition of $Y^{1}_{C^{t}(1)}$ and Equations \eqref{eq:prp1:delta:bound:pre}, it follows that
	\begin{equation}\label{eq:prop1:union}
	\begin{aligned}
	&P\left(\cup_{1\leq t\leq \tT}Y^{1}_{C^{t}(1)}\leq\maxmu\right)
	\\&= P\left(\cup_{1\leq t\leq \tT}V^{1}_{C^{t}(1)}\leq\maxmu-\FUB(C^{t}(1))\right)
	\\&\leq \sum_{t=1}^{\tT}\sum_{t=1}^{\tT}P\left(V^{1}_{N}
	\leq\maxmu-\FUB(N)\right)
	\leq \delta^{\alpha}e^{-\alpha\G/3}.
	\end{aligned}
	\end{equation}
	Recall that $T$ stands for the random final time step. Now, we assume that the algorithm terminated at a time step larger than $\tT$, namely, $T>\tT$ (we consider the case of $T\leq\tT$ later). We denote the first time step at which an arm has been sampled for $\frac{\tT}{|\KK|}$ times by $\tS$ and we note that $\tS\leq\tT$. By the fact that for every time step $Y^{k^{*}}_{C^{t}(k^{*})}\geq Y^{1}_{C^{t}(1)}$ holds for the chosen arm $k^{*}$ it follows that
	\begin{equation}\label{eq:prop1:union2}
	V^{k^{*}}_{C^{\tS}(k^{*})}+\FUB(C^{\tS}(k^{*}))=Y^{k^{*}}_{C^{\tS}(k^{*})}\geq Y^{1}_{C^{\tS}(1)}\, .
	\end{equation}
	Then, by Equation \eqref{time:bound} it follows that 
	\begin{equation}\label{eq:tT:lower:bound}
	|\KK|^{\alpha-1}\left(\frac{\G-\ln(\delta)}{\FF\left(\epsilon\right)}\right)^{\alpha}\leq \frac{1}{|\KK|}e^{\frac{\alpha \G}{3}}= \frac{\tT}{|\KK|}\, .
	\end{equation}
	Now, by Equation \ref{eq:tT:lower:bound}, the fact that $C^{\tS}(k^{*})=\frac{\tT}{|\KK|}$ and the increasing of $\FF^{-1}$ (and the decreasing of $\FUB$) it is obtained that
	$$\FUB(C^{\tS}(k^{*}))\leq \epsilon'\, ,$$
	where $\epsilon'=\FF^{-1}\left(\left(|\KK|\left(\G-\ln(\delta)\right)\right)^{1-\alpha}\left(\FF\left(\epsilon\right)\right)^{\alpha}\right)$.
	Therefore, by Equation \eqref{eq:prop1:union} the $(\epsilon',\delta')$-correctness with $\delta'=\delta^{\alpha}$ and $\epsilon'=\FF^{-1}\left(\left(|\KK|\left(\G-\ln(\delta)\right)\right)^{1-\alpha}\left(\FF\left(\epsilon\right)\right)^{\alpha}\right)$ is obtained for the case of $T>\tT$.
	
	For the case of $T\leq\tT$, by the fact that the algorithm terminated and the condition in step \ref{cond2} it follows that $\FUB(C^{T}(k^{*}))< \epsilon\leq\epsilon'$. Then, since 
	\begin{equation*}
		V^{k^{*}}_{C^{T}(k^{*})}+\FUB(C^{T}(k^{*}))=Y^{k^{*}}_{C^{T}(k^{*})}\geq Y^{1}_{C^{T}(1)}\,,
	\end{equation*}
	by Equation \eqref{eq:prop1:union} the $(\epsilon',\delta')$-correctness is obtained for the case of $T\leq\tT$.
	\ignore{
	Then, by the fact that $C^{\tS}(k^{*})=\frac{\tT}{|\KK|}$ and by Equation \eqref{time:bound} it follows that 
	$$|\KK|^{\alpha-1}\left(\frac{\G-\ln(\delta)}{\FF\left(\epsilon\right)}\right)^{\alpha}\leq \frac{1}{|\KK|}e^{\frac{\alpha \G}{3}}< \frac{\tT}{|\KK|}\, ,$$it is obtained by the increasing of $\FF^{-1}$ (and the decreasing of $\FUB$) that
	$$\FUB(\frac{\tT}{|\KK|})\geq \FF^{-1}\left(\left(|\KK|\left(\G-\ln(\delta)\right)\right)^{1-\alpha}\left(\FF\left(\epsilon\right)\right)^{\alpha}\right)\, .$$
	Therefore, by Equation \eqref{eq:prop1:union} the $(\epsilon',\delta')$-correctness with $\delta'=\delta^{\alpha}$ and $\epsilon'=\FF^{-1}\left(\left(|\KK|\left(\G-\ln(\delta)\right)\right)^{1-\alpha}\left(\FF\left(\epsilon\right)\right)^{\alpha}\right)$ is obtained.
}

	Now we continue with analyzing the sample complexity. First we recall that $T$ stands for the random final step. In additional, by the same considerations as in the proof of Theorem \ref{thm:alg1} it follows by Equation \eqref{time:bound} that $T\leq e^{L/3}$. For the case in which
	\begin{equation*}
	\overline{E}_{1}\triangleq\bigcap_{1\leq t<T}\left\{\max_{k\in\KK}Y^{k}_{C^{t}(k)}\geq\maxmu-\epsilon'\right\},
	\end{equation*}
	where $\epsilon'$ is defined above, the necessary condition for sampling from arm $k$,
	\begin{equation*}
	Y^{k}_{C^{k}(1)}\geq Y^{1}_{C^{t}(1)},
	\end{equation*}
	occurs only when the event
	\begin{equation*}
	\overline{E}^{k}_{2}(t)\triangleq\left\{\mu^{*}_{k}+\FUB\left(C^{t}(k)\right)\geq \maxmu-\epsilon'\right\},
	\end{equation*}
	occurs. But
	\begin{equation*}
	\overline{E}^{k}_{2}(t)\subseteq \left\{C^{t}(k)\leq \frac{\G-\ln(\delta)}{\FF\left(\maxmu-\epsilon'-\mu^{*}_{k}\right)}\right\}.
	\end{equation*}
	Therefore, it is obtained that
	\begin{equation}\label{prop:bound_set_N}
	C^{T}(k)\leq \max\left(\lfloor\frac{\G-\ln(\delta)}{\FF\left(\maxmu-\epsilon'-\mu^{*}_{k}\right)}\rfloor+1,N_0\right).
	\end{equation}
	Now, recall the sets $M\left(\epsilon\right)$ and $N\left(\epsilon\right)$ defined in the proof of Theorem \ref{thm:alg1}.
	
	By using the bound in Equation \eqref{C_bound} for the arms in the set $M(\epsilon')$, the bound in Equation \eqref{prop:bound_set_N} for the arms in the set $N(\epsilon')$ and the bound in Equation \eqref{time:bound}, it is obtained that
	\begin{equation}\label{eq:prop:final:1}
	E[T]\leq \left(1-P\left(\overline{E}_{1}\right)\right)e^{\frac{\G}{3}}+P\left(\overline{E}_{1}\right)\overline{\Phi}\left(\epsilon\right),
	\end{equation}
	where
	\begin{equation*}
	\begin{aligned}
	\overline{\Phi}\left(\epsilon\right)\triangleq&\sum_{k\in N(\epsilon')}\left(\lfloor\frac{\G-\ln(\delta)}{\FF\left(\min\left(\epsilon_0,\maxmu-\epsilon'-\mu^{*}_{k}\right)\right)}\rfloor+1\right)
	\\&+\sum_{k\in M(\epsilon')}\left(\lfloor\frac{\G-\ln(\delta)}{\FF\left(\epsilon\right)}\rfloor+1\right).
	\end{aligned}
	\end{equation*}
	In addition, by Equations \eqref{eq:prop1:union} and \eqref{eq:prop1:union2}, it follows that
	\begin{equation}\label{eq:prop:final:2}
	\begin{aligned}
	P\left(\overline{E}_{1}\right)&\geq1-P\left(\cup_{1\leq t\leq \tT}Y^{1}_{C^{t}(1)}\leq\maxmu\right)
	\\&\geq 1-\delta^{\alpha}e^{-\alpha\G/3}.
	\end{aligned}
	\end{equation}
	Furthermore, by the definitions of the sets $N(\epsilon')$ and $M(\epsilon')$, it can be obtained that
	\begin{equation}\label{eq:prop:final:3}
	\overline{\Phi}\left(\epsilon'\right)\leq\sum_{k\in \KK}\lfloor\frac{\G-\ln(\delta)}{\overline{\Theta}_k}\rfloor+1.
	\end{equation}
	where $\overline{\Theta}_k=\min\left\{\max\left(\epsilon,\maxmu-\epsilon'-\mu^{*}_{k}\right),\epsilon_0\right\}$.
	Therefore, by Equation \eqref{eq:final:2}, \eqref{eq:prop:final:1} and \eqref{eq:prop:final:3} and the fact that $\overline{\Phi}\left(\epsilon\right)\leq e^{\frac{\G}{3}}$ the bound on the sample complexity is obtained.
	\qed

\subsection{Proof of Proposition \ref{prop:2}}\label{proof:prop:2}
\proof
	Here, by Equation \eqref{eq:prp1:delta:bound:pre} it follows that
	\begin{equation}\label{eq:prp2:delta:bound:pre}
	\begin{aligned}
	P\left(V^{1}_{N}\leq\maxmu-\FUB(N)\right)&\leq \delta'e^{-L},
	\end{aligned}
	\end{equation}
	where $\delta'=\delta^{\alpha} e^{-(\alpha-1)L}$. So, by Equation \eqref{eq:prp2:delta:bound:pre}, Equation \eqref{eq:delta:bound:pre} holds for $\delta'$. Then we can proceed exactly as in the proof of Theorem \ref{thm:alg1}, but with $\delta'$ and Equation \eqref{eq:prp2:delta:bound:pre} instead of $\delta$ and Equation \eqref{eq:delta:bound:pre}. Therefore the result is obtained.
	\qed

\subsection{Proof of Theorem \ref{thm:lower:single}}\label{app:proof:lower:sing}
\proof
Similarly to the proof of Theorem \ref{thm:lower:bound}, we begin by defining the following hypotheses $\{H_0,H_1\}$, where $F^{H_l}(\mu)$ stands for the CDF of the unified arm under hypothesis $H_l$ (where $l\in\{0,1\}$). Hypothesis $H_0$ is the true hypothesis, namely,
\begin{equation*}
H_{0}:\quad F^{H_{0}}(\mu)=F(\mu),
\end{equation*}
Under hypothesis $H_{1}$ the maximal value is $\mu^{*,H_1}=\maxmu+\epsilon$, and Assumption \ref{assumption:1} still holds. Here, similarly to the proof of Theorem \ref{thm:lower:bound} we define $P^{\epsilon}\triangleq 1-\frac{1}{|\KK|}\FF\left(\epsilon_0\right)+\frac{1}{|\KK|}\FF\left(\maxmu+\epsilon-\mu^{*}_{k}\right)$ and let $$\overline{\mu}=\sup_{\mu\leq \maxmu}\{\mu|F(\mu)\leq P^{\epsilon}\}\,.$$ Then we set
\begin{equation*}
\begin{aligned} F^{H_{1}}(\mu)=&\gamma F(\mu)\boldsymbol{1}_{(-\infty,\overline{\mu}_{k})}(\mu)
\\&+\left(F(\mu)+\left(\gamma-1\right)F(\overline{\mu})\right)\boldsymbol{1}_{[\overline{\mu},\maxmu)}(\mu)
\\&+\left(1-\frac{1}{|\KK|}\FF\left(\maxmu+\epsilon-\mu\right)\right)\boldsymbol{1}_{\left[\maxmu,\maxmu+\epsilon\right]}(\mu)
\\&+\boldsymbol{1}_{(\maxmu+\epsilon,\infty)}
\end{aligned}
\end{equation*}
where $\gamma=1-\frac{\FF\left(\epsilon\right)}{|\KK|F(\overline{\mu})}$ and $\boldsymbol{1}_\Theta$ stand for the indicator function of the set $\Theta$.

Here, for showing that Assumption \ref{assumption:1} holds under hypothesis $H_1$ we need to show that,
$$1-F^{H_1}(\maxmu+\epsilon-\epsilon')\geq\frac{1}{|\KK|}\FF(\epsilon')$$
for every $0\leq\epsilon'\leq\epsilon_0$. Therefore, by the same considerations as in the proof of Lemma \ref{lem:assum} it follows that Assumption \ref{assumption:1} holds under hypothesis $H_1$.

If hypothesis $H_{1}$ is true, the algorithm should provide a reward greater than $\maxmu$. We use $E_{l}$ and $P_{l}$ (where $l\in\{0,1\}$) to denote the expectation and probability respectively, under the
algorithm being considered and under hypothesis $H_{l}$.
Now, let
\begin{equation*}
t'=\frac{1}{16\gamma'}\ln\left(\frac{3}{20\delta}\right),
\end{equation*}
where $\gamma'=\frac{\FF(\epsilon)}{|\KK|}$.

Recall that $T$ stands for the total number of samples from the arm. Now, we assume we run an algorithm which is $(\epsilon,\delta)$-correct under $H_{0}$ and that $E_{0}[T]\leq t$ for this algorithm. We will show that this algorithm cannot be $(\epsilon,\delta)$-correct under hypothesis $H_{1}$. Therefore, an $(\epsilon,\delta)$-correct algorithm must have $E_{0}[T]>t$.

Define the following events:
\begin{itemize}
	\item
$A=\{T\leq 4t\}$. By the same consideration as in the proof of Theorem \ref{thm:lower:bound} (for the events $\{A_{k}\}_{k\in\KK}$), it follows that if $E_{0}[T]\leq t$, then $P_{0}(A)\geq\frac{3}{4}$.
\item
Let $B$ stand for the event under which the chosen sample is strictly above $\maxmu$, and $B^{C}$ for its complementary (the chosen sample is smaller or equal to $\maxmu$). Clearly, $P_{0}\left(B^{C}\right)=1$.
\item
We define the event $C$ to be the event under which all the samples obtained from the unified arm are on the interval $[-\infty,\maxmu]$. Clearly, $P_{0}(C)=1$.
\item
We let $D$ denote the event under which for any number of samples $t\leq4t'$ from the unified arm, the number of samples which are on the interval $(-\infty,\overline{\mu}]$ is bounded, namely,
$$D\triangleq\left\{\max_{1\leq t\leq 4t'}\sum_{i=1}^{t}x_i-tF\left(\overline{\mu}\right)<15t'F\left(\overline{\mu}\right)\right\}$$
where $x_i$ is a R.V which equals to $1$ if the $i$-th sample is on the interval and $0$ otherwise. Now, by the same consideration as in the proof of Theorem \ref{thm:lower:bound}, it follows that $P_{0}\left(D\right)\geq\frac{9}{10}$.
\end{itemize}

Define now the intersection event $S=A\cap B^{C}\cap C\cap D$. We have shown that $P_{0}(A)\geq\frac{3}{4}$, $P_{0}(B)=1$, $P_{0}(C)=1$ and $P_{0}(D)=\frac{9}{10}$, from which it is obtained that $P_{0}\left(S\right)\geq\frac{13}{20}$.

Now, let $h$ to be the history of the process. Then by the same considerations as in the proof of Theorem \ref{thm:lower:bound}, for $\alpha=\frac{\gamma'}{F(\overline{\mu})}$, it follows that,
$$\frac{dP^{H}_{k}}{dP^{H}_{0}}(h)\Indi{S_{k}}\geq\left(1-\alpha\right)^{\frac{1}{\alpha}\ln\left(\frac{3}{20\delta}\right)}\Indi{S_{k}}\,.$$ Therefore we have the following,
\begin{align*}
P_{1}\left(B^{C}\right) &\geq P_{1}\left(S\right)=E_{0}\left[\frac{dP_{1}}{dP_{0}}(h)\Indi{S}\right]
\\
&\geq E_{0}\left[\left(1-\alpha\right)^{\frac{1}{\alpha}\ln\left(\frac{3}{20\delta}\right)}\Indi{S}\right]
\\&\geq\left(1-\alpha\right)^{\frac{1}{\alpha}\ln\left(\frac{3}{20\delta}\right)}P_{0}\left(\Indi{S}\right)
\\&> \frac{3}{20}e^{-\ln\frac{3}{20\delta}}\geq\delta\,,
\end{align*}
where in the last inequality we used the facts that $\left(1-\epsilon\right)^{\frac{1}{\epsilon}}\geq e^{-1}$.

We found that if an algorithm is $(\epsilon,\delta)$-correct under hypothesis $H_{0}$ and $E_{0}[T]\leq t$, then, under hypothesis $H_{1}$ this algorithm returns a sample that is smaller by at least $\epsilon$ than the maximal possible reward with a probability of $\delta$ or more, hence the algorithm is not $(\epsilon,\delta)$-correct.
Therefore, any $(\epsilon,\delta)$-correct algorithm, must satisfy $E_{0}[T]> t$. Hence the lower bound is obtained.

\qed

\subsection{Proof of Theorem \ref{thm:alg:Single}}\label{app:proof:of:al:2}
\proof
Since sampling from the unified arm consists of choosing one arm out of the $|\KK|$ arms (with equal probability), and then, sampling from this arm, it follows that, $F\left(\maxmu-\epsilon\right)\leq\left(1-\frac{\FF(\epsilon)}{|\KK|}\right)$. Also, we note that $(1-\epsilon)^\frac{1}{\epsilon}\leq e^{-1}$ for every $\epsilon\in(0,1]$. Therefore, for $N=\lceil\frac{-\ln(\delta)|\KK|}{\FF(\epsilon)}\rceil+1$,
\begin{equation}
P\left(V^{1}_{N}<\maxmu-\epsilon\right)=\left(F\left(\maxmu-\epsilon\right)\right)^{N}\leq \left(1-\frac{\FF(\epsilon}{|\KK|}\right)^{N}< \delta,
\end{equation}
where $V^{1}_{N}$ is the largest reward observed among the first $N$ samples. Hence, the algorithm is $(\epsilon,\delta)$-correct. The bound on the sample complexity is immediate from the definition of the algorithm.

\qed

\end{document}